# A New High-performance Entity Extraction Model Based on Machine Reading Comprehension

Xiaobo Jiang, *Member, IEEE*, Kun He, Jiajun He, and Guangyu Yan

*Abstract*—**Entity extraction is a critical technique in natural language processing to capture structured information from massive texts. Related models based on Machine Reading Comprehension (MRC) have more advantages than other mainstream methods. However, the lack of focus and in-depth interaction between question and context, which is not in line with human reading norms, limits the comprehension ability of these models, and the entity nesting problem brings about the omission or misjudgment of the answers (i.e., target entities). Therefore, this work presents a new and effective MRC-based entity extraction model—MRC-I2DP. It uses the proposed gated interactive attention mechanism to adjust the importance of each text part to the answers, then calculates multi-level interactive attention for question and context to enhance the understanding and recognition of target entities. It also exploits a designed 2D prob-encoding module, TALU function, and mask mechanism to strengthen the detection for all possible boundaries of target entities to improve the recall and precision of extraction. Experiments show that MRC-I2DP outperforms most state-of-the-art models on seven related datasets from the scientific and public domain, achieving a performance boost of 2.1% ~ 10.4% over baseline models in the F1 score.**

## I. INTRODUCTION

E NTITY extraction is a critical technique in natural language processing to capture structured knowledge and information (including terms in a specific domain like Protein, Material, Algorithm, and key phrases such as Task, Method, Processing, etc.) from massive texts. Previous related models mainly use sequence labeling [1]-[8] or span enumeration and classification [9]-[12] to judge whether it is an entity or not. The accuracy and efficiency are relatively low.

In contrast, the model based on Machine Reading Comprehension (MRC) method for entity extraction, which transforms the target entity type into questions and regards the entities as answers, is more consistent with human reading and searching for the required information and has more advantages over other mainstream methods. First, asking the model questions with entity categories is much more efficient to capture corresponding entities specifically rather than uniformly labeling all entities and classifying them one by one. Secondly, MRC or question answering (QA) is designed to improve the model's understanding of text content, helps entity extraction to get more accurate results based on comprehension.

In addition, by locating the answers directly instead of tagging them verbatim, the probability of long entities with more information disrupted inside is relieved, which improves the quality of extraction.

However, there are still two significant problems in the relevant entity extraction models based on QA or MRC.

On the one hand, there is a lack of focus in the text during the reading stage (not all parts are related to the target entity) and a lack of in-depth interaction between question and context, which is not in line with the standards of human reading comprehension. We usually pay attention to the paragraph whose content is closest to the given question, fully connect (i.e., interact) the contextual information to better understand the context, and then locate the possible answers. However, the existing related models only concatenate the paragraph and question to perform shallow unified processing [13]-[16] or merely increase the number of questions [17], so that the model can only give the direct answer after a simple extensive reading, which limits its comprehension ability. Therefore, when extracting *Task*, *Method*, and the other type of phrasal entities with a higher understanding requirement in scientific literature, as shown in Fig. 1(a), the performance is poor.

On the other hand, the problem of incomplete or inaccurate positioning of the answers (i.e., target entities) at the extraction stage, resulting in omission or misjudgment. These missing entities sometimes are essential, and misjudged ones bring us wrong information. Therefore, fully and accurately positioning all the answers is as important as the reading process. However, relevant models get the answer positions by using the output 1D prob-distribution [13], [14], [17] or generating and matching two independent start and end position distributions of the answer [15], [16]. The former cannot fully extract nested entities with intensive information, as shown in Fig. 1(b), and the latter is prone to entity misjudgment.

Aiming at the above problems of related works, an MRC-based entity extraction model, MRC-I2DP is constructed in this paper, which is more effective and more in line with human reading comprehension. By exploiting the proposed gated interactive attention mechanism and 2D probability encoding, the model enhances the entity understanding and positioning respectively, thus improving the overall extraction performance. Concretely, the main contributions of this work are as follows:

- A gated interactive attention mechanism is proposed, guided by the norms of human reading comprehension. Firstly, an adaptive gating matrix used in the proposal adjusts the importance of each text part to the answers to provide the focus for the model. Secondly, the multi-level

*(Corresponding author: Kun He.)*
The authors are with the School of Electronic and Information Engineering, South China University of Technology, Guangzhou 510610, China (e-mail: jiangxb@scut.edu.cn;  hk15616172426@163.com;  hjjun1996@163.com; 13650300559@163.com).

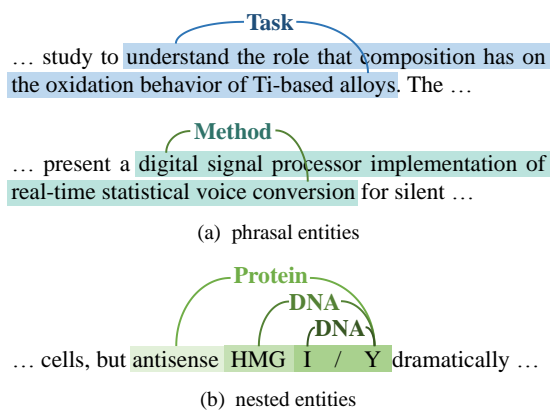

Fig. 1. Examples of phrasal entities with higher understanding requirement in scientific literatures and nested entities in information-intensive texts.

interactive attention calculation is applied to the context and the question to make the model deeply think with the target entity type, enhancing the overall attention on aimed entities and alleviating their semantic interweaving with other words. As a result, the comprehension and recognition ability of the model improved.

- A 2D prob-encoding module for answer positioning is proposed, equipped with a new probability function named TaLU and a 2D distribution mask mechanism. The module constructs the 2D position coordinate matrix of the answer, strengthening the boundary detection of all possible entities (especially nested entities) and increasing the extraction recall. Besides, each element in the square matrix uniquely determines an accurate start and end position pair of the potential entity, which avoids mismatching in two independent 1D distributions and improves the extraction accuracy.

- An MRC-I2DP model is proposed based on the above gated interactive attention mechanism and 2D prob-encoding to effectively extract entities from massive texts in the scientific and public domain. The model uses parallel computing to avoid a dramatic increase in time complexity and takes a masking mechanism to filter much noise to prevent the accuracy decline caused by the imbalance between positive and negative samples.

Experiments show that MRC-I2DP achieved F1 scores of 80.2%, 91.4%, 85.0%, 70.8%, 62.6%, 88.2% and 88.5% on 5 scientific domain literature datasets: GENIA [18], ADE [19], SOFC [4], SciERC [20], ScienceIE [21] and 2 public domain datasets: ACE2004 [22], ACE2005 [23], which outperforms most state-of-the-art models, achieving a performance boost of 2.1% to 10.4% over baseline models, proved to be of strong entity extraction ability and generalization.

## II. RELATED WORK

### A. Mainstream Methods for Entity Extraction

Entity extraction, first proposed at the *Sixth Message Understanding Conference* (MUC-6), is a critical technique in natural language processing for automatic information extraction and the basis of other tasks like reasoning and prediction. With the development of deep learning, the model architecture for entity extraction has shifted from the initial rule-and-feature-based construction to the neural network framework. The scope of application is also expanding, covering public domain texts and documents in various scientific fields.

Collobert et al. [1] proposed the first entity extraction model based on neural network, which combines Convolution Neural Network (CNN) with Conditional Random Field (CRF) to extract a string of words tagged as the same entity type by sequence labeling. This idea of word-by-word tagging is still the most popular entity extraction method in current applications. For example, Strakov'a et al. [2] combined the Bidirectional Long Short-Term Memory (BiLSTM) network, CRF, and pre-trained BERT model to extract *Organization*, *Geopolitics,* and other types of entities from ACE2004 and ACE2005 datasets. Jiang et al. [3] also used the architecture of BERT+BiLSTM to extract *Protein*, *DNA*, *RNA*, and others from GENIA. In addition, Friedrich et al. [4] capture *Materials*, *Devices,* and other types of entities from the SOFC dataset by a combined model of the pre-trained SciBERT, BiLSTM, and CRF. And Sahrawat et al. [5] used similar model structures to extract *Task*, *Processing* and other phrasal entities from the ScienceIE dataset.

In recent years, some entity extraction models have been improved based on sequential tagging to better deal with the more complex and diversified knowledge and information. For example, Shibuya and Hovy [6] proposed a Path-BERT model, which regards the tagged sequence as a second-best path, extracting entities iteratively in an outside-to-inside way. Fu et al. [7] proposed a new tagging algorithm named TreeCRF, which extracts entities by modeling a tree structure of their tags. And Kong et al. [8] further incorporated the graph-structured information during tagging.

Another kind of mainstream entity extraction method is span-based classification. All the spans in the text are enumerated and classified to determine whether it is the target entity. For example, Eberts and Ulges [9] proposed a BERT-based span classification model named SpERT to extract entities from ADE and SciERC. Wang et al. [10] further combine CNN on this framework to obtain span representation with local semantic information. Zhong and Chen [11] proposed a span classification model based on SciBERT, which pre-filtered spans with lengths larger than the setting threshold, and Shen et al. [12] added a target detection algorithm called SoftNMS based on span classification to enhance the detection of spans that contain target entities.

Although the above mainstream related models achieved some success in the scientific and public domain, the extraction by tagging word-by-word or enumerating span does not accord with the purposeful reading and information search, leading to a lack of the pertinence to target entity type and the understanding of text content. They also have disadvantages of high time complexity and/or introduce much noise during training, which results in low efficiency. Therefore, under the guidance of human reading and information search norms, this article uses the form of reading comprehension for entity extraction, transforms the tagging and the enumeration into the way of questioning model to improve the comprehension and extraction efficiency.

### B. QA/MRC-based Entity Extraction

Compared with the widely used sequence tagging or span classification methods, the entity extraction model based on QA or MRC acts more like human beings to read and search for the required information and has more advantages. Concretely, the question provides the model with a priori message of the target entity category, which is beneficial for extracting corresponding entities pertinently. Secondly, entity extraction based on QA or MRC helps the model obtain more accurate results after understanding the text content. In addition, directly locating the answer rather than tagging word-by-word improves the efficiency and reduces the probability that long entities with much information break from the inside during extraction.

In the past two years, some researchers have applied the QA or MRC form to entity extraction by transforming target entity types into questions and regarding the entity as the answer to the question, to a certain extent, enhancing the model's comprehension and improving performance. For example, the team from Shannon Technologies [13] proposed a BERT-based Multi-turn QA model to transform the extraction of different types of entities into a multi-round Q&A process and locate the target entity through a BMEO (Begin, Middle, End, and Outside) fully connected layer. Jain et al. [14] also proposed a QA-based framework for entity extraction, which obtains the location distribution of entities through a layer of 1D convolution. Li et al. [15] combined the BERT with the MRC task form, previously designed questions corresponding to each entity type, and got the starting and ending positions of answers through a layer of linear transformation after splicing the text and question representation at the output. Zheng et al. [16] and Zhao et al. [17] used the same model architecture for entity extraction, increased the number of fully connected layers and questions, respectively.

Although the above entity extraction models adopt the task form of QA or MRC, they still lack the focus in the text and ignore the further interaction between questions and contextual information, which is inconsistent with human reading habits. The shallow unified processing or a just numerical increase of questions limits the model's ability to understand. Besides, their way of answer positioning is prone to the omission or misjudgment of target entities. In this paper, MRC-I2DP uses the proposed gated interactive attention mechanism to provide the focus for the model while calculating the multi-level interactive attention between contextual information and question to enhance the comprehension and recognition ability, as well as constructing a 2D coordinate probability matrix to improve the recall and precision of extraction.

### III. MRC-I2DP MODEL

The MRC-I2DP proposed in this paper is a new and efficient model for automatic entity extraction, suitable for scientific literature and public domain texts. The training and inference process is in the task form of MRC so that the model requires relative questions or query statements designed for all target entity types in advance. For example, when extracting Neural Network type entities from AI domain literature, the query can be "Please find the neural networks such as CNN, LSTM, GRU or Transformer, BERT, etc."

The input of the model is a concatenation of text and a query sentence. Since the role of the query is to provide prepared information of the target entity type, we suggest only keeping the corresponding keywords rather than paying attention to the whole semantic coherence. Thus the above query can be further simplified to "neural network CNN LSTM GRU Transformer BERT" to reduce the length of the input sequence and the amount of computation.

The output of the model consists of three parts: two vector distributions and a probability distribution matrix. The former two indicate the start and end positions of the target entities, while the last is the crucial 2D entity coordinates. Each element in the vector distribution means "the probability of the index predicted as the initial start or end position." and each one in the 2D probability distribution matrix refers to "the probability of the horizontal and vertical indexes of the coordinates being predicted as the start and end position, respectively." The three probability distributions above and the corresponding mask mechanism with a probability threshold $T$ helps the model to obtain the final start and end positions to extract entities from the text.

The MRC-I2DP model takes the proposed 2D probability encoder as the core module, combined with a gated interactive attention mechanism that conforms to the human reading comprehension norms to enhance the understanding and attention of the target entities and the detection of all possible boundaries. In addition, Byte Pair Encoding (BPE) for tokenization and DeBERTa [24] for pre-trained word vectors are used to improve the entity extraction performance of the model. The overall architecture of our MRC-I2DP, as shown in Fig. 2, consists of three main modules.

- The Pre-trained module: conducts a feature extraction of the input after BPE tokenization by deep self-attention to obtain the representation of word vectors with the global semantic information.
- Pointer module: implements the preliminary interactive attention calculation on the output representation of the pre-trained module above. The result is mapped to a vector distribution with the same length as the input and then normalized to probability.
- 2DP-Encoder module: proceeds a further gated interactive attention calculation on context and question from output representation of pre-trained module. The result is mapped to a square matrix of 2D probability distribution with the same length and width as the input and then normalized to probability.

### A. The Pre-trained Module

The pre-trained module aims to get the word representations (i.e., vectors) that are injected with the global semantic information, to enhance the model's ability to understand the content of the text.

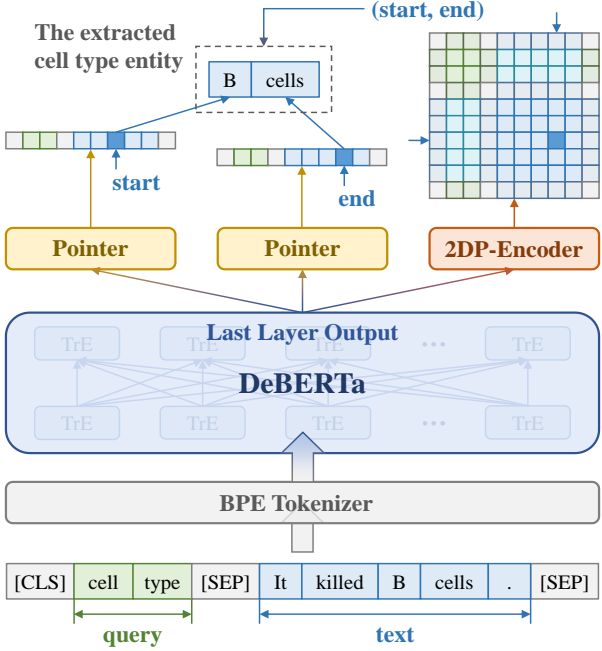

Fig. 2. Overall architecture of the MRC-I2DP model.

First of all, the input sentence of the text and designed query are concatenated to an input format accepted by the pre-trained module, as shown in Eq. (1).

$$x_{in} = \{[CLS], q_0, q_1, q_2, ...., [SEP], t_0, t_1, t_2, ...., [SEP]\} \quad (1)$$

where the "$[CLS]$" token represents the semantic information of the whole sentence, $\{q_0, q_1, q_2, \ldots\}$ and $\{t_0, t_1, t_2, \ldots\}$ are the query of target entity type and the text sentence, respectively; and the "$[SEP]$" token is used as the separator.

Secondly, to reduce the number of rare words in the text and alleviate the Out-Of-Vocabulary (OOV) problem, the model takes BPE encoding (essentially a subword algorithm based on the frequency statistics of the continuous byte pair) for tokenization. The rare words are decomposed into common affixes and fragments to improve the model's cognitive ability to these obscure domain conceptual term entities.

The input of the pre-trained module is the sentence that went through word tokenization. There are two main reasons why to choose DeBERTa as the pre-trained module. For one, the self-attention score of an entity depends not only on its content but also on the relative position between words. For example, the dependence of the tokens "deep" and "learning" when they appear adjacent is much stronger than that when they are far away, which is particularly prominent in scientific literature. DeBERTa adopts a modified self-attention mechanism based on the separation of content and position representation, which improves the effectiveness of the self-attention scores of entities. For another, an Enhanced Mask Decoder (EMD) in the output layer of DeBERTa alleviates the mismatch between pre-trained and fine-tuning to some extent, improving the robustness of the model for entity extraction in different domain texts.

In this paper, we selected the representation of the last hidden layer in DeBERTa as the output $\boldsymbol{h_o}$ of this module:

$$\boldsymbol{h_o} = deberta\big(bpe(x_{in})\big)[-1] \in \square^{\ l \times d} \quad (2)$$

where $l$ is the sequence length of $\boldsymbol{h_o}$ after BPE tokenization and $d$ is the dimension of word representation.

### B. Pointer Module

The Pointer module is responsible for mapping the word representation $\boldsymbol{h_o}$ into a vector of probability distribution with the length of $l$ to obtain the preliminary prediction result of the start or end positions of the target entities. Different from the simple linear mapping in the related work of Li et al. [15] and Zheng et al. [16], Pointer exploit the vector representation of "$[CLS]$" which contains global information to calculate the further interactive attention between text, query, and contextual information to enhance the concern to each whole target entity. As a result, the extraction performance of some phrasal entities with higher contextual comprehension requirements in the scientific literature is improving. The algorithm of the Pointer module shows in Fig. 3.

The input representation $\boldsymbol{h_o}$ first goes through a layer of the linear transformation, in which the activation function is an approximate of Gaussian Error Linear Unit (GELU) with the expression of Eq. (3), to obtain a new representation $\boldsymbol{h}$ as shown in Eq. (4).

$$gelu(x) = 0.5x \cdot [1 + tanh(0.8x + 0.036x^3)] \quad (3)$$

$$\boldsymbol{h} = gelu(\boldsymbol{W} \cdot \boldsymbol{h_o} + \boldsymbol{b_w}) \in \square^{\ l \times d} \quad (4)$$

Then $\boldsymbol{h}$ interacts with the global semantic vector $\boldsymbol{h_{cls}}$ to get the vector of candidate probability distribution $\boldsymbol{p_h}$.

$$\boldsymbol{p_h} = talu(\boldsymbol{h} \cdot \boldsymbol{h_{cls}}) \in \square^{\ l} \quad (5)$$

TaLU is a proposed probabilistic (i.e., normalized) function whose expression shows in Eq. (6). It maps the calculated interactive attention score to the range of (0,1). Since the derivative range of the TaLU function is 4.0 times that of the widely used sigmoid function, it can effectively improve the differentiation of the probability at different positions, which is helpful to the boundary detection of the model.

$$talu(x) = e^x / (e^x + e^{-x}) \in (0,1) \quad (6)$$

In addition, the input $\boldsymbol{h_o}$ also goes through a layer of linear compression ($d \rightarrow 1$) to obtain another vector of the candidate probability distribution $\boldsymbol{p_c}$.

$$\boldsymbol{p_c} = talu(\boldsymbol{v}^T \cdot \boldsymbol{h_o} + b_v) \in \square^{\ l} \quad (7)$$

The initial probability distribution $\boldsymbol{s}$ (or $\boldsymbol{e}$) of the start (or end) position as the final output is the mean of $\boldsymbol{p_h}$ and $\boldsymbol{p_c}$.

### C. 2DP-Encoder Module

The 2DP-Encoder module is devoted to further improves the model's extraction performance for some nested entities with intensive information, as well as some phrasal entities that require contextual understanding in the scientific literature.

First of all, there are two cases of entity nesting. One is the nesting in different entity types. For example, "SiC power device" is a *Device* type entity in which the nested "SiC" is a *Material* type entity. Another is the nesting in the same entity types, such as that one start position corresponds to multiple end positions, or one end position corresponds to several start positions. For example, "convolution neural network" and its nested "convolution" are both *Algorithm* type entities with the same start; "HMG I/Y" and its nested "I/Y" are both *DNA* type entities with the same end. These two kinds of overlap are common in various domain texts, but existing related models are unable or hard to deal with them. However, 2DP-Encoder furthest enhances boundary detection for all possible nested entities in different or same type by mapping the input $\boldsymbol{h_o}$ into a 2D probability distribution matrix with length and width of both $l$, to improve the recall of nested entity extraction.

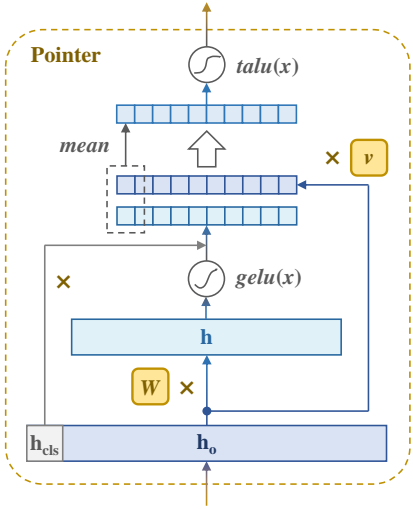

Fig. 3. Algorithm of the Pointer module.

Secondly, the module also combines a gated interactive attention mechanism that accords with the specification of human reading comprehension proposed in this paper. By adaptively adjusting the importance of each part of the text to the answers (i.e., target entities) through the gating matrix, and fully interacting with the global semantic information, the entity type information in the query, and the contextual information of the text, the model's ability to understand the specified entity type and the overall attention to the target entities further enhance. The algorithm of the 2DP-Encoder module shows in Fig. 4.

On the one hand, the input $\boldsymbol{h_o}$ goes through two independent linear transformations, respectively. Two representations: $\boldsymbol{h_g}$ and $\boldsymbol{h_{2D}}$, used to calculate the gating matrix and 2D probability distribution matrix, are obtained.

$$\boldsymbol{h_g} = gelu(\boldsymbol{W_g} \cdot \boldsymbol{h_o} + \boldsymbol{b_g}) \in \square^{l \times d} \quad (8)$$

$$\boldsymbol{h_{2D}} = gelu(\boldsymbol{W_{2D}} \cdot \boldsymbol{h_o} + \boldsymbol{b_{2D}}) \in \square^{l \times d} \quad (9)$$

Then the $\boldsymbol{h_g}$ is mapped to the range of (0,1) after normalization, and the gated weight matrix $\boldsymbol{g}$ is obtained.

$$\boldsymbol{g} = 1 / (1 + e^{-\boldsymbol{h_g}}) \in \square^{l \times d} \quad (10)$$

The gating matrix $\boldsymbol{g}$ is employed to automatically adjust the importance of different text parts related to target entities for the $\boldsymbol{h_o}$ and $\boldsymbol{h_{2D}}$. The result is a vector representation named $\boldsymbol{h_f}$ as Eq. (11) shows, in which the "$\odot$" represents element-wise multiplication.

$$\boldsymbol{h_f} = \boldsymbol{g} \odot \boldsymbol{h_o} + (1 - \boldsymbol{g}) \odot \boldsymbol{h_{2D}} \in \square^{l \times d} \quad (11)$$

The following step is the calculation of interactive attention between $\boldsymbol{h_f}$ and $\boldsymbol{h_{2D}}$ to get the first candidate 2D probability distribution matrix $\boldsymbol{m_{sp}}$ of the position coordinates.

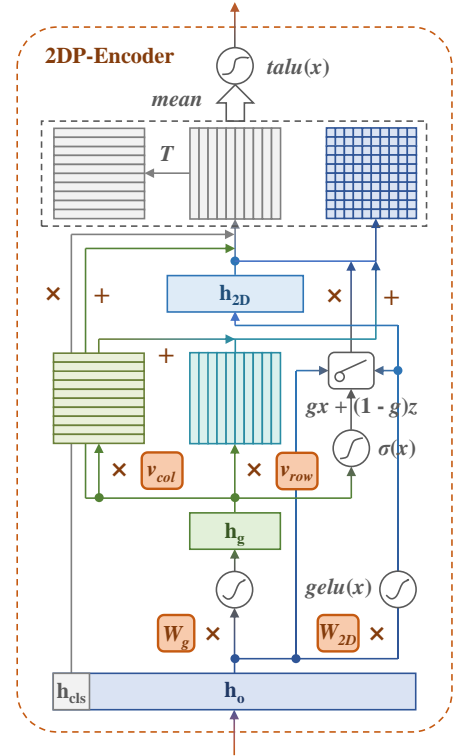

Fig. 4. Algorithm of the 2DP-Encoder module.

$$\boldsymbol{m_{sp}} = talu(\boldsymbol{h_{2D}} \cdot \boldsymbol{h_f^T}) \in \square^{l \times l} \quad (12)$$

On the other hand, the vector representation $\boldsymbol{h_{cls}}$ with global semantic also interacts with $\boldsymbol{h_g}$ and $\boldsymbol{h_{2D}}$. The resulting vector distribution $\boldsymbol{m_h}$ is expanded and superimposed by rows and columns to generate the second candidate 2D probability distribution matrix $\boldsymbol{M_h}$.

$$\boldsymbol{m_h} = talu[(\boldsymbol{h_g} + \boldsymbol{h_{2D}}) \cdot \boldsymbol{h_{cls}}] \in \square^{l} \to \boldsymbol{M_h} \in \square^{l \times l} \quad (13)$$

To further enrich the information in the 2D probability distribution and achieve more accurate positions of target entities, 2DP-Encoder also considers the distribution information $\boldsymbol{M_{row}}$ and $\boldsymbol{M_{col}}$ in the direction of separate rows and columns. The calculation expressions show in equations (14) and (15), respectively.

$$\boldsymbol{m_{row}} = talu(\boldsymbol{h_g} \cdot \boldsymbol{v_{row}^T} + b_{row}) \in \square^{l} \to \boldsymbol{M_{row}} \in \square^{l \times l} \quad (14)$$

$$\boldsymbol{m_{col}} = talu(\boldsymbol{h_g} \cdot \boldsymbol{v_{col}^T} + b_{col}) \in \square^{l} \to \boldsymbol{M_{col}} \in \square^{l \times l} \quad (15)$$

The final output 2D probability distribution $\boldsymbol{m}$ is the mean of $\boldsymbol{m_{sp}}$, $\boldsymbol{M_h}$, $\boldsymbol{M_{row}}$, and $\boldsymbol{M_{col}}$. In addition, to obtain more accurate results of the start and end positions of the entities, $\boldsymbol{m}$ should be used jointly with the two vector distributions $\boldsymbol{s}$ and $\boldsymbol{e}$ generated by the Pointer module and a proposed corresponding 2D mask mechanism.

### D. Training of The Model

To avoid the introduction of noise samples and make the extraction results more accurate. The proposed MRC-I2DP model should integrate with the masking mechanism shown in Fig.5 during training.

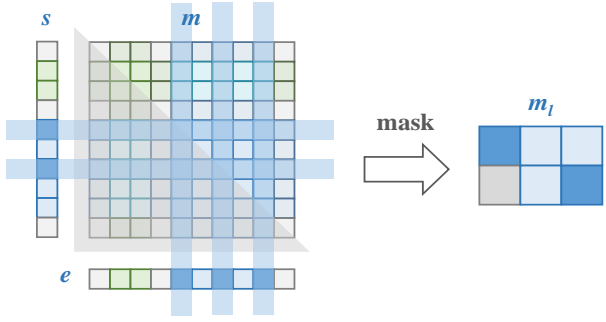

Fig. 5. Algorithm of the 2DP-Encoder module.

On the one hand, since the model takes BPE word tokenization, some complete obscure words in the input sentence are split into multiple segments. The elements of probability distribution with the index corresponding to the middle segments should be zero as they can be neither the starting positions nor the ending positions of the candidate entities. Besides, since the index of the starting positions cannot be larger than that of the ending ones, the lower triangular elements of the generated 2D probability distribution matrix $\boldsymbol{m}$ need to be zeroed.

On the other hand, to avoid the performance degradation caused by the imbalance between positive and negative samples, the probability threshold of training is set to $T_t$. We selected the elements greater than $T_t$ from $\boldsymbol{s}$ and $\boldsymbol{e}$. Their corresponding indexes are as the preliminary starting and ending positions of all possible entities, respectively.

$$p_s = \{i \mid \boldsymbol{s}[i] > T_t, 0 \le i < l\} \qquad (16)$$

$$p_e = \{j \mid \boldsymbol{e}[j] > T_t, 0 \le j < l\} \qquad (17)$$

Then only a few elements in $\boldsymbol{m}$ whose row indexes belong to $p_s$ and column index belong to $p_e$ are screened out as the final masked result $ml$ for the coordinates of the extracted entities.

$$m_l = \{\boldsymbol{m}[i][j] \mid i \in p_s, j \in p_e\} \qquad (18)$$

The overall masking mechanism shows in Fig. 5.

The loss function for training consists of three weighted parts: the loss of $\boldsymbol{s}$ and real start position distribution $\boldsymbol{y}_s$, the loss of $\boldsymbol{e}$ and real ending position distribution $\boldsymbol{y}_e$, as well as the loss of masked 2D probability matrix $\boldsymbol{m}_l$ and real position coordinate matrix $\boldsymbol{y}_m$.

$$loss = \frac{1-\lambda}{2}\big[f(\boldsymbol{s}, \boldsymbol{y}_s) + f(\boldsymbol{e}, \boldsymbol{y}_e)\big] + \lambda \cdot f(\boldsymbol{m}_l, \boldsymbol{y}_m) \qquad (19)$$

The function $f(\boldsymbol{x}, \boldsymbol{y})$ represents the Binary Cross Entropy (BCE) between predictions and true labels. The expression of BCE shows in Eq. (20).

$$f(\boldsymbol{x}, \boldsymbol{y}) = -\frac{1}{n}\sum_{i=0}^{n-1}[y_i \ln x_i + (1 - y_i)\ln(1 - x_i)] \qquad (20)$$

$\lambda$ in Eq. (19) is an adjustable parameter, which represents the proportion of 2D probability distribution in the $loss$.

### E. Evaluating of The Model

In the evaluation stage, we added the length restriction of the target entities on the mask used in training. Thus the elements in 2D probability distribution whose vertical coordinate (i.e., the end position) exceed the horizontal coordinate (i.e., the start position) by more than a maximum entity length of $l_m$ are set to zero. Making sure some entities with extra length are filtered to enhance the robustness of the model for answer positioning. The final result $p$ of the 2D coordinates shows below:

$$p = \{(i, j) \mid \boldsymbol{m}_l[i][j] > T_e, j - i \le l_m\} \qquad (21)$$

By using all the start and end positions in $p$, the corresponding text content can be extracted from the input sentence and decoded (by BPE) as the final target entities.

## IV. EXPERIMENT AND DISCUSSION

### A. Datasets and Evaluation Metric

In this paper, we choose five scientific literature datasets and two public domain datasets which are most representative and widely used in entity extraction research, to verify the entity extraction ability and generalization of the model, as shown in Table Ⅰ. These datasets include:

- **GENIA**[1]: The widely used GENIAcorpus3.02p selected in this paper consists of 2000 abstracts of biomedical literature in *PubMed* database, including 36 types of entities, such as *Protein*, *DNA*, *RNA*, *Cell Type*, *Cell Line*, etc. The preprocessing steps are consistent with Jiang et al. [3], Fu et al. [7], and most of the related work, that is, only to consider the five entity types listed above, and the first 90% of sentences are used for training while the remaining 10% for testing.

- **ADE**[2]: The dataset derived from the literature abstracts in *PubMed* and related electronic treatment records consists of 4272 sentences, including two entity types: *Drugs* and *Adverse-Effect*. We take the same 10-fold cross-validation as Eberts and Ulges [9] for training.

- **SciERC**[3]: It is composed of 500 abstracts in AI conference literature with 2687 sentences in total, of which the test set accounts for 551 items. The dataset includes six entity types, concretely, *Task*, *Method*, *Metric*, *Material*, *Generic*, and *Other-Scientific-Term* (OST).

- **ScienceIE**[4]: It is proposed by the *SemEval 2017 Task 10*. This dataset consists of 500 literature paragraphs in terms of computer science, material science, and physics in the *Science Director* database, of which the test set accounts for 100. It includes three types of entities: *Task*, *Material*, and *Processing*.

- **SOFC**[5]: It consists of 45 carefully selected literature about solid oxide fuel cells in material science, of which 34 are training sets and 11 are test sets, containing 7630 and 1836 sentences, respectively. There are four entity types in this dataset: *Material*, *Device*, *Experimental Operation*, and *Experimental Value*.

- **ACE2004**[6] and **ACE2005**[7]: Two widely used public domain datasets derived from news texts. The entity types

---

[1] http://www.nactem.ac.uk/genia/genia-corpus
[2] https://sites.google.com/site/adecorpus/
[3] http://nlp.cs.washington.edu/sciIE/
[4] https://scienceie.github.io/resources.html
[5] https://github.com/boschresearch/sofc-exp_textmining_resources
[6] https://catalog.ldc.upenn.edu/LDC2005T09
[7] https://catalog.ldc.upenn.edu/LDC2006T06

TABLE I
OVERVIEW OF DADASETS FOR EXPERIMENT

| Datasets | Source Texts | Entity Types | Designed Query Sentences |
|---|---|---|---|
| **GENIA** | abstracts of the biomedical literature | *Protein* | protein nitrogenous organic compounds body tissues muscle hair collagen enzymes antibodies |
| | | *DNA* | DNA deoxyribonucleic acid |
| | | *RNA* | RNA ribonucleic acid |
| | | *Cell Type* | cell type |
| | | *Cell Line* | cell line |
| **ADE** | medical literature abstracts and treatment records | *Drug* | drug interferon methotrexate alpha beta lithium acid amiodarone |
| | | *Adverse-Effect* | severe acute syndrome symptoms reaction effects toxicity disease |
| **SciERC** | abstracts of the AI conference literature | *Task* | task processing image speech video information translation classification recognition |
| | | *Method* | method techniques approach algorithm model framework network |
| | | *Metric* | metrics accuracy precision recall f1 bleu rouge robustness |
| | | *Material* | data corpus corpora text image speech video |
| | | *Generic* | general common scientific term |
| | | *OST* | other scientific term |
| **ScienceIE** | literature paragraphs about computer science, material and physics | *Task* | task analysis problems design |
| | | *Material* | material data particles surface |
| | | *Processing* | process model method algorithm approach |
| **SOFC** | the literature in the field of material science | *Material* | material anode cathode electrolyte fuel interlayer support |
| | | *Device* | device |
| | | *Experiment* | experiment evoking word |
| | | *Value* | value voltage current power resistance thickness temperature |
| **ACE2004 & ACE2005** | news texts | *PER* | person human single individual group |
| | | *LOC* | geographical location areas landmass mountain water river |
| | | *ORG* | organization company corporation agency institution group |
| | | *FAC* | facility building man-made structure airport highway bridge |
| | | *VEH* | vehicle move carry transport helicopter train ship motorcycle |
| | | *WEA* | weapon physical device instrument harming gun gunpowder |
| | | *GPE* | geographical political country nation region city state government |

include seven categories: *Person* (*PER*), *Location* (*LOC*), *Organization* (*ORG*), *Facility* (*FAC*), *Weapon* (*WEA*), *Vehicle* (*VEH*), and *Geographical Entities* (*GPE*).

We make an exploration of the above datasets, including the proportion of nested entities, whether it contains phrasal entities with higher comprehension requirements, as well as the shortest, the longest, and the average length of the entities, as shown in Table II, for the analysis of experimental results.

The evaluation metric adopts the commonly used *Precision* (*P*), *Recall* (*R*), and *F1* score in related works. The calculation expressions show below.

$$P = \frac{the\ number\ of\ correct\ entities\ extracted}{the\ total\ number\ of\ entities\ extracted} \quad (22)$$

$$R = \frac{the\ number\ of\ correct\ entities\ extracted}{the\ actual\ total\ number\ of\ entities} \quad (23)$$

$$F1 = \frac{2 \times P \times R}{P + R} \quad (24)$$

To be consistent with other related works, we calculate the F1 score on the ADE and the SOFC datasets by inter-class averaging (Macro) while calculating the results on others by total averaging (Micro).

### B. Experimental Equipment and Parameters

We carried out all the experiments in this work on a single 11G GTX 1080 Ti. The sentence length of all datasets is limited to 64 words after tokenization. The training takes the AdamW as an optimizer, and the corresponding weight decay is 0.01. The training probability threshold $T_t$ of the mask is 0.5, the dropout rate is 0.3, and the weight $\lambda$ of the 2D prob-distribution matrix is 0.1. Other parameters are adjusted for experiments on different datasets, as shown in Table III.

TABLE II
DATA EXPLORATION OF THE EXPERIMENTAL DATASETS

| Datasets | Entity Length | | | Entity Nesting Rate | Including Phrasal Entities |
|---|---|---|---|---|---|
| | Min | Max | Avg | | |
| **GENIA** | 1 | 43 | 4.1 | 10.0% | No |
| **ADE** | 1 | 17 | 2.7 | 2.3% | No |
| **SciERC** | 1 | 29 | 4.2 | 3.4% | Yes |
| **ScienceIE** | 1 | 51 | 4.6 | 20.2% | Yes |
| **SOFC** | 1 | 36 | 3.9 | 0.0% | Yes |
| **ACE2004** | 1 | 67 | 3.2 | 24.0% | No |
| **ACE2005** | 1 | 64 | 2.9 | 22.0% | No |

In addition, except that the experiments on ADE datasets use pre-trained word vectors of PubMedBERT [25] specific for biomedical fields, DeBERTa is used for other datasets.

### C. Comparative Experiments and Discussion

First, comparative experiments are carried out on seven scientific and public domain datasets for entity extraction to verify the overall performance and generalization of the proposed MRC-I2DP model. The results show in Table IV (for scientific domain) and Table V (for public domain). All the

TABLE III
PARAMETERS FOR EXPERIMENTS

| Datasets | Training Epochs | Batch Size | | Learning Rate | $T_e$ | $l_m$ |
|----------|-----------------|------------|-----|---------------|-------|-------|
| | | Train | Test | | | |
| **GENIA** | 5 | 4 | 16 | 5e-6 | 0.50 | — |
| **ADE** | 10 | 16 | 32 | 3e-5 | 0.60 | 10 |
| **SciERC** | 10 | 4 | 16 | 5e-6 | 0.46 | 20 |
| **ScienceIE** | 15 | 4 | 16 | 5e-6 | 0.52 | — |
| **SOFC** | 20 | 4 | 16 | 5e-6 | 0.72 | 32 |
| **ACE2004** **ACE2005** | 10 | 4 | 16 | 5e-6 | 0.50 | — |

TABLE IV
COMPARATIVE EXPERIMENTS IN SCIENTIFIC DOMAIN

| Datasets | Models | P | R | F1 |
|----------|--------|------|------|------|
| **GENIA** | *BERT + BiLSTM [3] | 76.7 | 76.7 | 76.8 |
| | Multi-agent Comm [26] | 77.2 | 76.6 | 76.9 |
| | TCSF [27] | 78.2 | 76.5 | 77.3 |
| | Path-BERT [6] | 77.8 | 76.9 | 77.4 |
| | BERT-Seq2Seq [2] | — | — | 78.2 |
| | BioBERT + TreeCRFs [7] | 78.2 | 78.2 | 78.2 |
| | BERT + BENSC [28] | 79.2 | 77.4 | 78.3 |
| | **MRC-I2DP (ours)** | **80.1** | **80.2** | **80.2** |
| **ADE** | DAPNA [8] | 90.8 | 86.2 | 88.4 |
| | *SpERT [9] | 89.0 | 89.6 | 89.3 |
| | CMAN [29] | — | — | 89.4 |
| | BERT + FFNN [3] | — | — | 89.6 |
| | BERT + TSE [30] | — | — | 89.7 |
| | TriMF [31] | 89.5 | 91.3 | 90.4 |
| | **MRC-I2DP (ours)** | **91.3** | **91.5** | **91.4** |
| **SciERC** | *BERT + BiLSTM [32] | — | — | 63.8 |
| | BERT-MRC [15] | 69.5 | 62.2 | 65.6 |
| | SPE [10] | — | — | 66.9 |
| | SpERT [9] | 68.5 | 66.7 | 67.6 |
| | ENPAR [33] | — | — | 67.9 |
| | PURE [11] | — | — | 68.9 |
| | **MRC-I2DP (ours)** | **71.1** | **70.5** | **70.8** |
| **ScienceIE** | *BERT + BiLSTM [5] | — | — | 52.2 |
| | SciBERT + JLSD [34] | — | — | 54.6 |
| | BERT + JLSD [34] | — | — | 55.4 |
| | BERT-MRC [15] | 57.5 | 54.2 | 55.8 |
| | Span + RoBERTa + CRF [35] | — | — | 58.6 |
| | Span + XLNet + CRF [35] | — | — | 60.1 |
| | **MRC-I2DP (ours)** | **66.1** | **59.3** | **62.6** |
| **SOFC** | *BERT + BiLSTM [4] | — | — | 79.7 |
| | SciBERT + BiLSTM [4] | — | — | 81.5 |
| | **MRC-I2DP (ours)** | **85.2** | **84.9** | **85.0** |

TABLE V
COMPARATIVE EXPERIMENTS IN PUBLIC DOMAIN

| Datasets | Models | P | R | F1 |
|----------|--------|------|------|------|
| **ACE2004** **(English)** | *Multi-turn QA [13] | 84.4 | 82.9 | 83.6 |
| | BERT-Seq2Seq [2] | — | — | 84.3 |
| | Path-BERT [6] | 85.9 | 85.7 | 85.8 |
| | BERT-MRC [15] | 85.1 | 86.3 | 86.0 |
| | BERT + TreeCRFs [7] | 86.7 | 86.5 | 86.6 |
| | BERT-Seq2Set [36] | **88.5** | 86.1 | 87.3 |
| | Span + BERT + SoftNMS [12] | 87.4 | 87.4 | 87.4 |
| | **MRC-I2DP (ours)** | 88.4 | **88.0** | **88.2** |
| **ACE2005** **(English)** | BERT-Seq2Seq [2] | — | — | 83.4 |
| | Path-BERT [6] | 83.8 | 84.9 | 84.3 |
| | *Multi-turn QA [13] | 84.7 | 84.9 | 84.8 |
| | BERT + TreeCRFs [7] | 84.5 | 86.4 | 85.4 |
| | MRC4ERE++ [17] | — | — | 85.5 |
| | BERT-MRC [15] | 87.2 | 86.6 | 86.9 |
| | Span + BERT + SoftNMS [12] | 87.4 | 87.4 | 86.7 |
| | BERT-Seq2Set [36] | 87.5 | 86.6 | 87.1 |
| | **MRC-I2DP (ours)** | **88.5** | **88.5** | **88.5** |

- **Multi-turn QA**: the first and most representative entity extraction model in the kind of works based on QA/MRC. It is widely used as a comparison in related researches and proposed by Shannon Technologies [13].

The experimental results in Table IV indicate that the entity extraction performance of the proposed MRC-I2DP model surpasses most of the representative and related SOTA models in the fields of biomedicine, material, AI, and computer science. Showing strong practicability and achieves a significant improvement in F1 compared with baselines by $2.1\% \sim 10.4\%$. As a whole, it fully proves that the model can be effectively applied to literature in the different scientific domains to realize automatic entity extraction.

The experimental results in Table V show that, for public domain news texts in which there are also a large number ($22.0\% \sim 24.0\%$) of nested entities in non-scientific fields, MRC-I2DP is also better than most representative and related SOTA models. It achieved an F1 increase of $3.7\% \sim 4.6\%$ relative to the baselines, proving that our model can also be effectively applied to public domain texts and has a strong generalization.

To help analyze the comprehension ability of MRC-I2DP to the target entities, a comparative histogram of the performance improvement (relative to baselines) in different kinds of scientific literature is constructed in this section, as shown in Fig. 6. It exhibits that the entity extraction performance of computer science and AI domain literature is generally lower than that of texts in other scientific fields. However, our MRC-I2DP model happens to have a significant F1 boost in these fields (up to $7.0\% \sim 10.4\%$). Given this result, we delivered the following explanation. Most of the knowledge and information in scientific fields like biomedicine and material science is domain conceptual term entities, such as *DNA*, *mRNA*, *ZnO*, etc. These noun entities contain neither words with semantic information nor some common roots or affixes, so they are relatively isolated and easy to identify in the text. But the entities in AI and computer science literature are mainly phrasal entities related to the research content, such listed comparisons are the representative works or the state-of-the-art (SOTA) related researches.

To make the comparison results more persuasive, we considered the models based on two kinds of mainstream entity extraction methods and the models based on the QA/MRC method. This work selected the most representative three described in the Related work section as the baseline models marked with "*" in Table IV and Table V. They are:

- **BERT + BiLSTM**: the most widely used mainstream model based on sequence tagging for entity extraction.
- **SpERT**: a mainstream entity extraction model based on span enumeration and classification, commonly used as the baseline model in related works, proposed by Eberts and Ulges [9].

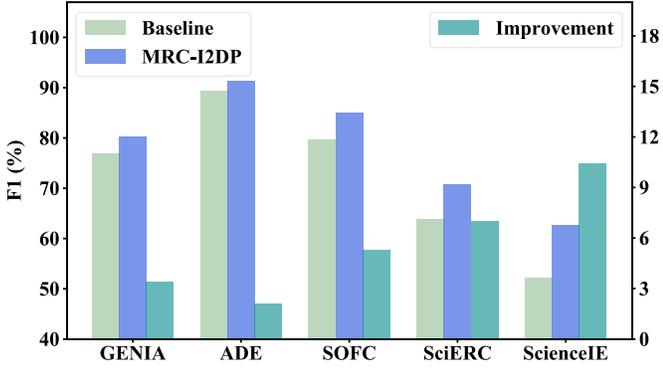

Fig. 6. Comparison of performance improvement (relative to baseline) on different scientific literature datasets.

| Datasets | Average Entity Length | Entity Nesting Rate | Including Phrasal Entities | Performance Improvement |
|---|---|---|---|---|
| **GENIA** | 4.1 | 10.0% | No | 3.4% |
| **ADE** | 2.7 | 2.3% | No | 2.1% |
| **SciERC** | 4.2 | 3.4% | Yes | 7.0% |
| **ScienceIE** | 4.6 | 20.2% | Yes | 10.4% |
| **SOFC** | 3.9 | 0.0% | Yes | 5.3% |
| **ACE2004** | 3.2 | 24.0% | No | 4.6% |
| **ACE2005** | 2.9 | 22.0% | No | 3.7% |

as *Task*, *Method*, *Process*, etc. They contain words with semantic information like "understanding", "combine", "split into", etc., which have higher semantic interweaving with other words in the text, thus brings difficulty for extraction.

The proposed MRC-I2DP model exploits the MRC task form and a gated interactive attention mechanism according to the norms of human reading comprehension. Enhancing the contextual understanding and the overall attention of the target entities, and to some extent, also reduces their semantic interweaving with other words. Therefore, both the above qualitative analysis and quantitative experimental results show that MRC-I2DP has a high ability to understand the content of the text and the target entities.

In addition, to further verify the model's ability to recognize phrasal entities with high comprehension needs and improve the recall when locating nested entities with dense information. The work constructs Table VI by using the average entity length, entity nesting rate, and whether the phrasal entities are included in each dataset in Table II as independent variables and takes the performance improvement on each baseline as dependent variables. We also draw the corresponding line charts, as shown in Fig. 7 and Fig. 8, respectively, for the observation and analysis of performance improvement on datasets with different average entity lengths and entity nesting rates. (Since it is difficult to carry out quantitative statistics for phrasal entities with high comprehension requirements, we adopted the average entity length as the indicator of the comprehension requirement for the model in this paper. The results of Fig. 1 and Table II also show that the average length of these phrasal entities is relatively longer.)

Fig. 7 illustrates that the overall trend in both scientific and public domain datasets is that "the longer the average length of the entities (i.e., the higher the comprehension requirement for the model), the greater the extraction performance improvement of the model." Fig. 8 also shows the general trend that "with the increase of entity nesting rate, the improvement of extraction performance also increases." It is further verified from two aspects that the model is powerful to understand and recognize the phrasal entities with higher comprehension needs and improving the recall when locating nested entities.

Besides, no matter in the results of Fig. 7 or Fig. 8, the model is best at dealing with the scientific literature that requires high comprehension ability, which is just the weakness of other existing entity extraction models. It sufficiently demonstrates the advantages of the proposed MRC-I2DP model over many other related works.

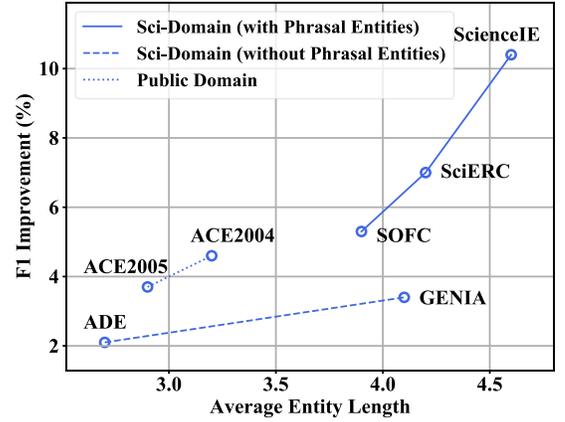

Fig. 7. Performance improvement with different average entity lengths.

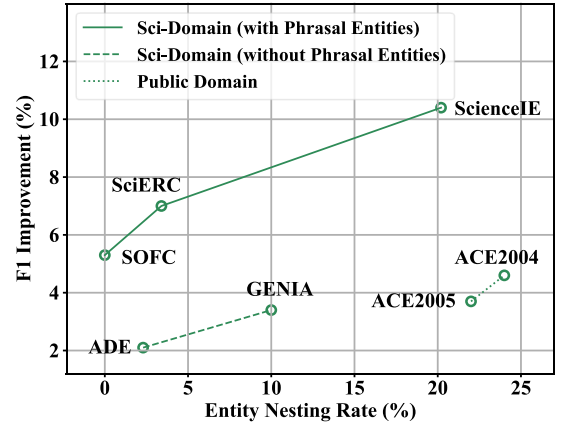

Fig. 8. Performance improvement with different entity nesting rate.

### D. Ablation Experiments and Discussion

The purpose of ablation experiments is to verify the effectiveness of the two critical algorithms proposed to improve the overall performance of entity extraction in MRC-I2DP, namely interactive attention mechanism, and 2D probability encoding. The interactive attention mechanism is used in the reading stage to enhance the understanding of the context with target entity type and the overall attention to target entities, to improve the recognition performance of phrasal entities, which require higher comprehension in the scientific literature. The 2D probability encoding module is mainly used in answer (i.e., target entity) positioning, aiming to fully and accurately extract the entities (especially those nested entities with intensive information), and further improve the quality and quantity of the extracted results.

Firstly, we carried out the ablation experiment of the interactive attention mechanism, removed the computing parts related to the interactive attention in two Pointer modules and 2DP-Encoder module, and modified the rest parts as same as Li et al. [15] do by concatenating text and question representations and putting them through a fully connected layer. Then re-trained and re-evaluated all the datasets. Secondly, we implemented the ablation of 2D probability encoding, removed the 2DP-Encoder module with the corresponding TaLU function and mask mechanism, and retained only two Pointer modules for generating the starting and ending position distributions as Zheng et al. [16] do to match the two 1D-distribution of start and end positions. Then conducted re-training and evaluating.

The ablation results show in Table Ⅶ, which display that the removal of the interactive attention mechanism results in a relative F1 decrease of 0.3% ~ 3.5% in entity extraction performance on each dataset, and the deletion of 2D probability encoding also relatively reduces the performance by 0.7% ~ 4.4%. As a whole, it proves the effectiveness of the two critical algorithms proposed.

For further exploring the function of interactive attention mechanism in enhancing model's ability of comprehension and phrasal entity recognition in scientific literature, in this section, we also take the average entity length as a degree quantification of the model's comprehension requirement. We construct the performance degradation line chart under the ablation of interactive attention mechanism, as shown in Fig. 9. The result exhibits a general trend that "the longer the average entity length (i.e., the higher the requirement for comprehension), the more the performance decreases." as the proposed interactive attention mechanism is more consistent with human reading comprehension. Firstly, it automatically adjusts the importance of each text part to the answers through an adaptive gating matrix, which is equivalent to the fact that we always decide which paragraphs are more important related to the question and then focus on them when reading. Secondly, the further interaction of entity type, contextual information, and text content is equivalent to our combination of question and context to deeply understand the relevant content, which is greatly helpful to the search of target entities. The ablation result proves the significance and effectiveness of the interactive attention mechanism in context understanding and phrasal entity extraction of scientific literature.

Besides, to further analyze the effectiveness of 2D probability encoding for answer positioning and nested entity extraction, in this section, we construct a performance degradation line chart corresponding to different entity nesting rates with the ablation of 2D probability encoding, as shown in intensive information and improve the extraction recall. So the higher the entity nesting rate, the greater the performance degradation. The ablation result proves the significance and effectiveness of 2D probability encoding in answer (entity) positioning and nested entity extraction.

TABLE Ⅶ
Ablation Experiments

| Datasets | Full Model | F1 and Relative Decrease | |
|---|---|---|---|
| | | - Iterative Attention | - 2D Probability Encoding |
| **GENIA** | 80.2 | 78.8 (-1.7%) | 78.6 (-2.0%) |
| **ADE** | 91.4 | 91.1 (-0.3%) | 90.8 (-0.7%) |
| **SciERC** | 70.8 | 69.2 (-2.3%) | 69.9 (-1.3%) |
| **ScienceIE** | 62.6 | 60.4 (-3.5%) | 60.7 (-3.1%) |
| **SOFC** | 85.0 | 84.1 (-1.1%) | 84.0 (-1.2%) |
| **ACE2004** | 88.2 | 87.7 (-0.6%) | 84.3 (-4.4%) |
| **ACE2005** | 88.5 | 87.9 (-0.7%) | 85.2 (-3.7%) |

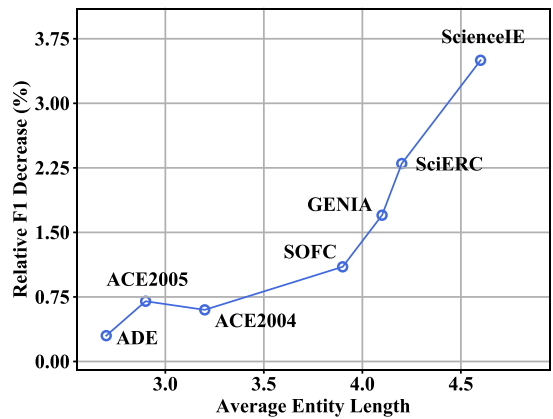

Fig. 9. Relative performance degradation with different average entity lengths after removing interactive attention mechanism.

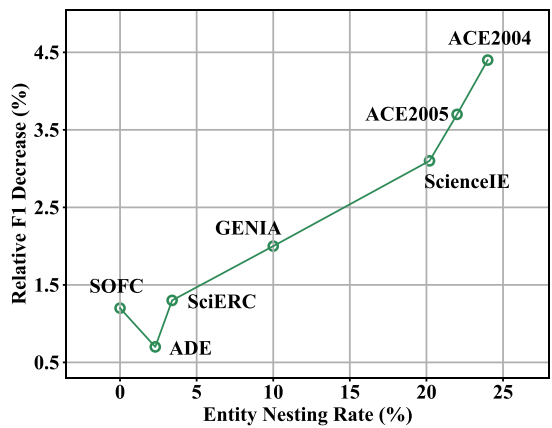

Fig. 10. Relative performance degradation with different entity nesting rates after removing 2D probability encoding.

Fig. 10. The result delivers an overall trend of "the higher the nesting rate of entities, the more performance degradation." The reason that, compared with the positioning method used in related works (i.e., to generate a unique 1D entity range distribution for labeling or to generate and match two independent starting and ending position distributions), the proposed 2D probability encoding has an enhanced ability to detect all the entity boundaries. Each element in the generated coordinate matrix uniquely determines the start-end position pair of an entity, which is of no mismatch error. Therefore, it can effectively deal with some nested entities that contain

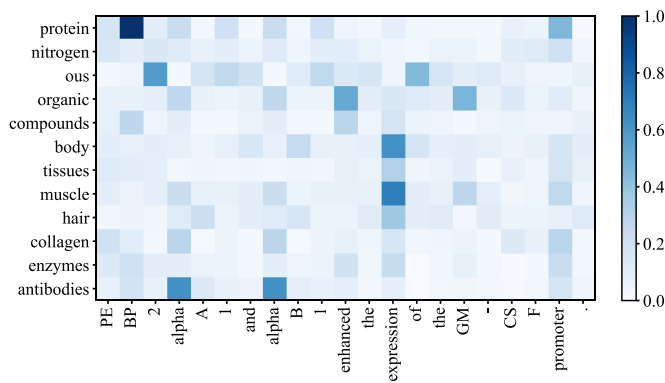

Fig. 11. Visualization of the interactive attention matrix.

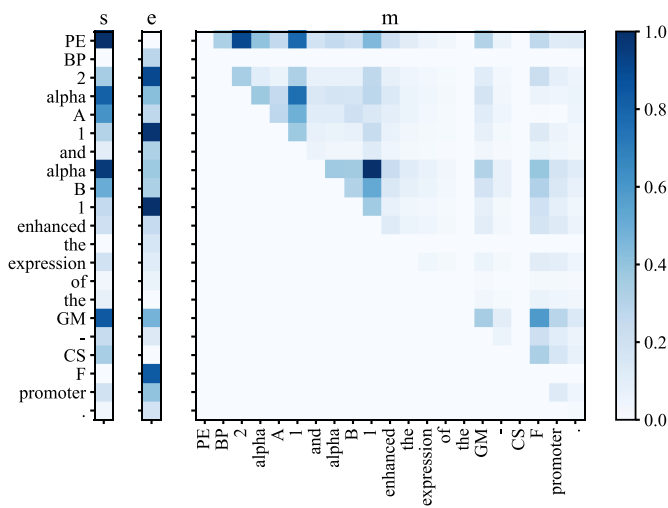

Fig. 12. Visualization of the 2D probability distribution matrix and the starting and ending position distributions.

### E. Visualization and Analysis

To show the performance of the MRC-I2DP model more clearly and intuitively, we selected one sample in preprocessed GENIA dataset in this section, of which the text sentence is "PEBP2 alpha A1 and alpha B1 enhanced the expression of the GM-CSF promoter." and the target entity type is *Protein*. The designed query sentence corresponding to this entity type is "protein nitrogenous organic compounds body tissues muscle hair collagen enzymes antibodies."

Firstly, the visualized interactive attention matrix in MRC-I2DP shows in Fig. 11. It illustrates that the word "PEBP2" has the highest attention score with the word "protein" in the matrix, which conforms to that PEBP2 is a protein. In addition, the attention between "alpha A1", "alpha B1", and "antibodies" and that between "expression", "body", and "muscle" are also high, better in line with the biomedical knowledge. Therefore, the model's ability to understand the context and the entity type is verified.

Secondly, we visualized start and end position distributions generated by the Pointer module for target entities and the 2D probability distribution generated by 2DP-Encoder. As shown in Fig. 12, the start position distribution $s$ provides four words in the head of predicted protein entities, namely "PE", "alpha", "alpha", and "GM". The end position distribution $e$ also delivers four tail words: "2", "1", "1", and "F". For most of the relevant models, only the results of start and end position distribution (i.e., $s$ and $e$, respectively) are used for matching the range of target entities, so that in this case, just "PEBP2", "alpha A1", "alpha B1" and "GM-CSF" can be obtained. However, the output distribution $m$ of 2D probability encoding proposed in this article can capture five targets, including the above four entities and an officially labeled nested entity "PEBP2 alpha A1" with the type of *Protein*. The visualization result demonstrates the effectiveness of the proposed 2D probability encoding for answer positioning and nested entity extraction.

## V. CONCLUSION

In this paper, we proposed a new high-performance model named MRC-I2DP for entity extraction. Through the MRC task form, equipped with the gated interactive attention mechanism and 2D prob-encoding, the model enhanced comprehension of the context and the target entities and improved the ability to capture the answers fully and accurately. Therefore, it effectively alleviates the problem of related models with poor behavior in extracting the phrasal entities in scientific literature and information-intensive nested entities.

The comparative experiments on both scientific and public domain datasets show that the MRC-I2DP outperforms most related representative and SOTA works. It achieves a significant F1 boost of up to 10.4% compared with baseline models, showing higher extraction ability and generalization. Ablation experiments and visualization further confirm that the interactive attention mechanism and 2D prob-encoding can effectively handle the phrasal entities with high comprehension needs and information-intensive nested entities, thus improving the overall extraction quality and quantity.

However, we have only validated the proposed MRC-I2DP model on English texts, so the subsequent work is trying to extract entities from some related Chinese texts to verify its universality.

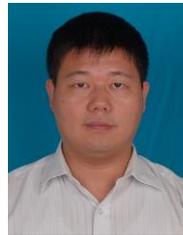

**Xiaobo Jiang** (Member, IEEE) received the B.S. and M.S. degree from Zhejiang University, Zhejiang, China, in 1994 and 1997, and the Ph.D. degree in Solid State Electronics and Micro-electronics, from the Chinese academy of sciences institute of microelectronics, Beijing, China, in 2004. After graduation, he joined South China University of Technology, Guangzhou, China. His research interests include artificial intelligence, natural language processing and artificial intelligence chip design.

**Kun He** received the B.S. degree from Hunan University, Hunan, China, in 2018, and is currently an undergraduate with the School of Electronic and Information Engineering of South China University of Technology. His research interests include natural language processing and deep learning.

**Jiajun He** received the B.S. degree and M.S. degree from South China University of Technology, Guangzhou, China, in 2018 and 2021. His research interests include natural language processing and speech recognition and processing.

**Guangyu Yan** received the B.S. degree from South China University of Technology, Guangzhou, China, in 2021, and is currently an undergraduate with the School of Electronic and Information Engineering of South China University of Technology. His research interests include natural language processing and machine learning.